\documentclass[letterpaper, 10 pt, conference]{ieeeconf}  

\IEEEoverridecommandlockouts                              

\usepackage{cite}
\usepackage{graphics} 
\usepackage{epsfig} 
\usepackage{mathptmx} 
\usepackage{times} 
\usepackage{amsmath} 
\usepackage{amssymb}  
\graphicspath{ {./images/} }
\usepackage{balance}
\usepackage{float}
\usepackage{textcomp}
\usepackage{algorithm,algpseudocode}
\usepackage{caption}
\usepackage{subcaption}

\usepackage{hyperref}
\newcommand{\dataacronym}{{DESK }}
\newcommand{\dataname}{{DESK }}

\title{\LARGE \bf
DESK: A Robotic Activity Dataset for Dexterous Surgical \\ Skills Transfer to Medical Robots
}

\author{Naveen Madapana$^{1 *}$, Md Masudur Rahman$^{2 *}$, Natalia Sanchez-Tamayo$^{1 *}$, Mythra V. Balakuntala$^{3}$,\\ Glebys Gonzalez$^{1}$, Jyothsna Padmakumar Bindu$^{3}$, L. N. Vishnunandan Venkatesh$^{3}$, Xingguang Zhang$^{1}$,\\ Juan Barragan Noguera$^{1}$, Thomas Low $^{4}$, Richard Voyles$^{3}$, Yexiang Xue$^{2}$, and Juan Wachs$^{1}$
\thanks{* These authors contribute equally to this work}%
\thanks{$^{1}$School  of  Industrial  Engineering, Purdue University, West Lafayette, IN, 47907, USA 
        {\tt\small nmadapan, sanch174, gonza337, zhan3275, barragan,jpwachs@purdue.edu}}
\thanks{$^{2}$ Department of Computer Science, Purdue University, West Lafayette, IN, 47907, USA 
{\tt\small rahman64, yexiang@purdue.edu}}
\thanks{$^{3}$College of Technology, Purdue University, West Lafayette, IN, 47907, USA
        {\tt\small mbalakun, jpadmaku, lvenkate, rvoyles@purdue.edu}}%
\thanks{$^{4}$ SRI International 
{\tt\small thomas.low@sri.com}}
}

\begin{document}
\maketitle
\thispagestyle{empty}
\pagestyle{empty}

\begin{abstract} 
Datasets are an essential component for training effective machine learning models. In particular, surgical robotic datasets have been key to many advances in semi-autonomous surgeries, skill assessment, and training. Simulated surgical environments can enhance the data collection process by making it faster, simpler and cheaper than real systems. In addition, combining data from multiple robotic domains can provide rich and diverse training data for transfer learning algorithms. In this paper, we present the DESK (Dexterous Surgical Skill) dataset. It comprises a set of surgical robotic skills collected during a surgical training task using three robotic platforms: the Taurus II robot, Taurus II simulated robot, and the YuMi robot. This dataset was used to test the idea of transferring knowledge across different domains (e.g. from Taurus to YuMi robot) for a surgical gesture classification task with seven gestures. We explored three different scenarios: 1) No transfer, 2) Transfer from simulated Taurus to real Taurus and 3) Transfer from Simulated Taurus to the YuMi robot. We conducted extensive experiments with three supervised learning models and provided baselines in each of these scenarios. Results show that using simulation data during training enhances the performance on the real robot where limited real data is available. In particular, we obtained an accuracy of 55\% on the real Taurus data using a model that is trained only on the simulator data. Furthermore, we  achieved an accuracy improvement of 34\% when 3\% of the real data is added into the training process.
\end{abstract}

\section{introduction}

Minimally invasive robotic surgery has evident advantages over traditional surgery, such as quick recovery, lower risks and lower catastrophic errors for patients, and thereby have become the standard of care for a wide variety of surgical procedures \cite{chellali_validation_2014}. However, these techniques require residents to spend a substantial amount of time practicing surgical maneuvers in simulation environments. In particular, the surgical robotic simulators play a crucial role in the training process of residents and novice surgeons, leading to significant improvement of their technical skills gradually and over time \cite{korets_validating_2011}. The tasks that are predominantly presented for training using the simulation and bench-top models include, but are not limited to, peg transfer, pattern cut, suture, and needle passing \cite{hung_comparative_2013}.

There is another key-benefit in using simulation: (a) It provides unlimited amount of training time while avoiding human/animal tissue manipulation or expensive single-use mock models \cite{peng_sim--real_2018} and (b) It allows researchers to collect enormous amount of data at a lower cost and in a scalable manner \cite{tobin_domain_2017}. While Da-Vinci datasets are publicly available including surgical maneuvers during a variety of procedures, there is a lack of datasets featuring other type of surgical robots which may be less popular. This limits the ability to generalize across other robotic platforms. This limitation is particularly critical when in-field deployable surgical robots are needed (e.g. military, disaster-relief scenarios). The scarcity of publicly available datasets prevents researchers from exploring novel strategies to transfer the knowledge gained in the surgical simulator to a variety of other robots, less common in the Operating Room (OR) but more suitable to field conditions.

The goal of this paper is to provide such dataset that can allow for "transfer learning" across robotic platforms. As a test-case study, we will rely on the peg transfer task, which is a task of primary importance in surgical skill learning. In this regard, this paper introduces a library of motions for the robotic peg transfer task obtained from multiple domains: robotic simulator and two real robots (Taurus and YuMi). We refer to this database as Dexterous Surgical Skills Transfer dataset (DESK). In addition to providing the dataset, we will present a proof-of-concept for learning to classify surgemes with a very limited or no training information from the real robots. The goal here is to transfer the knowledge gained from the simulator to accelerate the learning on real robots by leveraging on abundant data obtained from the simulation environment. This instance of transfer learning is known as domain adaptation, a scenario in transductive transfer learning \cite{pan_survey_2010}, where the source and target tasks are the same, but the data distribution of target and source domain are different. 

The main contributions of this work are to: 1) Provide an annotated dataset of surgical robotic skills (DESK) collected in three domains: two real robots with different morphology (Taurus II and YuMi) and one simulated robot (Taurus II), and 2) Transfer the knowledge gained from abundant simulation data to real robot's data in context of surgeme classification.

\section{background and related work}

The advancement of surgical robotic activity recognition and semi-autonomous surgeries benefits from the the availability of robotic surgical datasets. Two such prominent datasets are the JIGSAW \cite{gao_jhu-isi_2014} (JHU-ISI Gesture and Skill Assessment Working Set) and MISTIC-SL (Johns Hopkins Minimally Invasive Surgical Training and Innovation Center; Science of Learning Institute). These datasets comprise procedures preformed with the da Vinci Surgical System on a bench-top model, including synchronized video and kinematic data \cite{gao_unsupervised_2016}. A main advantage of these datasets is that they can allow elucidating patterns associated with skill learning. With this goal in mind, surgical tasks are decomposed into a finite set of maneuvers \cite{gao_jhu-isi_2014}. This process of decomposition is known as surgical skill modelling \cite{moustris_evolution_2011}. In this modelling technique, each surgical skill is represented as a sequence of atomic units referred as gestures or surgemes \cite{gao_jhu-isi_2014}. 


The MISTIC-SL dataset contains 49 right-handed trials of a suture throw followed by a surgeon's knot, employing in total four maneuvers: \textit{suture throw}, \textit{knot tying}, \textit{grasp pull run suture}, and \textit{intermaneuver segment} \cite{dipietro_recognizing_2016}. This dataset has been used to learn representations of surgical motions \cite{dipietro_unsupervised_2018} and recognize surgical activities \cite{dipietro_recognizing_2016}. However, the MISTIC-SL dataset is not publicly available at the moment. In contrast, the JIGSAW is a public dataset that contains 39 trials of \textit{suturing} task, 36 trials of \textit{knot tying}, and 28 trials of \textit{needle passing}. The data is annotated by one individual and manually segmented into 15 atomic units of intentional surgical activity called "gestures". This dataset has been used for multiple applications such as motion generation of expert demonstrations \cite{reiley_motion_2010}, recognition of surgical gestures \cite{ahmidi_dataset_2017}, and surgical trajectory segmentation \cite{krishnan_transition_2018}. Nonetheless, the data collection of the JIGSAW dataset did not intentionally introduce variability in the environment and initial conditions of the task. In realistic surgical tasks, the videos and kinematics will vary greatly between demonstrations stressing the need of surgical datasets that include variability in the experimental setup to facilitate generalization.

The segmentation and classification of time series data on surgical datasets has been evaluated on the JIGSAW and MISTIC-SL dataset using several approaches. Previous supervised learning methods include the use of hidden Markov models \cite{reiley_task_2009} \cite{tao_sparse_2012}, conditional random fields \cite{tao_surgical_2013}, and bag of spatio-temporal features \cite{ahmidi_dataset_2017}. More recent methods include the use of Recurrent Neural Networks for recognizing surgical activities \cite{dipietro_recognizing_2016, dipietro_unsupervised_2018}.
Nonetheless, these approaches were tested using data from the same distribution as the training data, and do not account for the disparity encountered from randomized initial conditions.

In surgical data analysis, many state-of-the-art techniques contributions are limited to study-specific data and validation metrics \cite{ahmidi_dataset_2017}. Recently, several efforts have been made in autonomous classification and execution of surgical tasks using multiple surgical robots and simulation platforms \cite{kehoe_autonomous_2014, schulman_case_2013,seita_fast_2018}. However, the obtained models are specific for a given platform and setup, and cannot be directly transferred to other robots/procedures. As surgical data sets become available, it is desirable to use principles of transfer learning to leverage previous models to accelerate learning in new domains, where it is essential to avoid collecting extensive data to retrain the models.

Research in transfer learning and domain adaptation has leveraged simulation environments to boost learning in real systems \cite{ bousmalis_using_2018,fang_multi-task_2018}. Knowledge transfer between a robot and a simulated environment is challenging due to the differences in data distribution. Simulated environments and real systems are often unalike in terms of interface used for controlling, object appearance, surface dynamics or even physical behaviour of the robots. Methods of domain randomization have been proposed to increase generalization in real scenarios for models trained in simulation \cite{james_transferring_2017, tobin_domain_2017}, by randomizing object position and appearance over the training set.

Transfer learning between dissimilar robots has been extensively studied in the area of Reinforcement Learning (RL)  \cite{barrett2010transfer,mahadevan1992enhancing,rusu2016sim,taylor2011integrating}. The work in \cite{devin2017learning} shows a modular policy strategy for networks that allows to jointly train data from different robots with a common task, or data coming for the same robot with varying tasks. Also, 
the work in \cite{malekzadeh_skills_2013} proposed to reproduce skills of an industrial robot on a flexible robotic prototype designed for surgical tasks. Their approach uses reward profiles learned from a demonstration step (source domain) to learn over the target domain in a RL setup.  In domain trasnfer scenarios, dimensionality reduction techniques can be used to transfer learning between different robots \cite{bocsi2013alignment}. Bócsi et al. proposed a knowledge transfer
approach that is agnostic to the robot parameters, were the source and the target dataset are reduced to a common lower-dimension manifold \cite{bocsi2013alignment}. This method does not require any kinematic knowledge of the target domain. Instead, the users must propose a bijective mapping between each dataset and the lower dimensional manifold. In contrast with this approach, our method does not require mapping back to the original space. We take a dimensionality reduction approach that intends to preserve the common features and the transfer learning is done directly in the reduced space. 

Our \dataacronym dataset provides RGB images, depth and kinematic information for the peg transfer task from multiple domains including two real robots (Taurus II and YuMi) and a simulation environment (Taurus II). These three robotic setups possess inherent variance in peg board configuration and object size and appearance. Additional variability is added to the dataset by randomizing the pick and place locations for the pegs and orientation of the board, while leaving the order of the pegs to be transferred unrestrained. In addition, the dataset contains examples of success and failure of surgemes employed during the task and subsequent recovery maneuvers.

\newpage

\section{\dataname dataset}

\begin{figure}[htpb]
\includegraphics[width = 1\columnwidth]{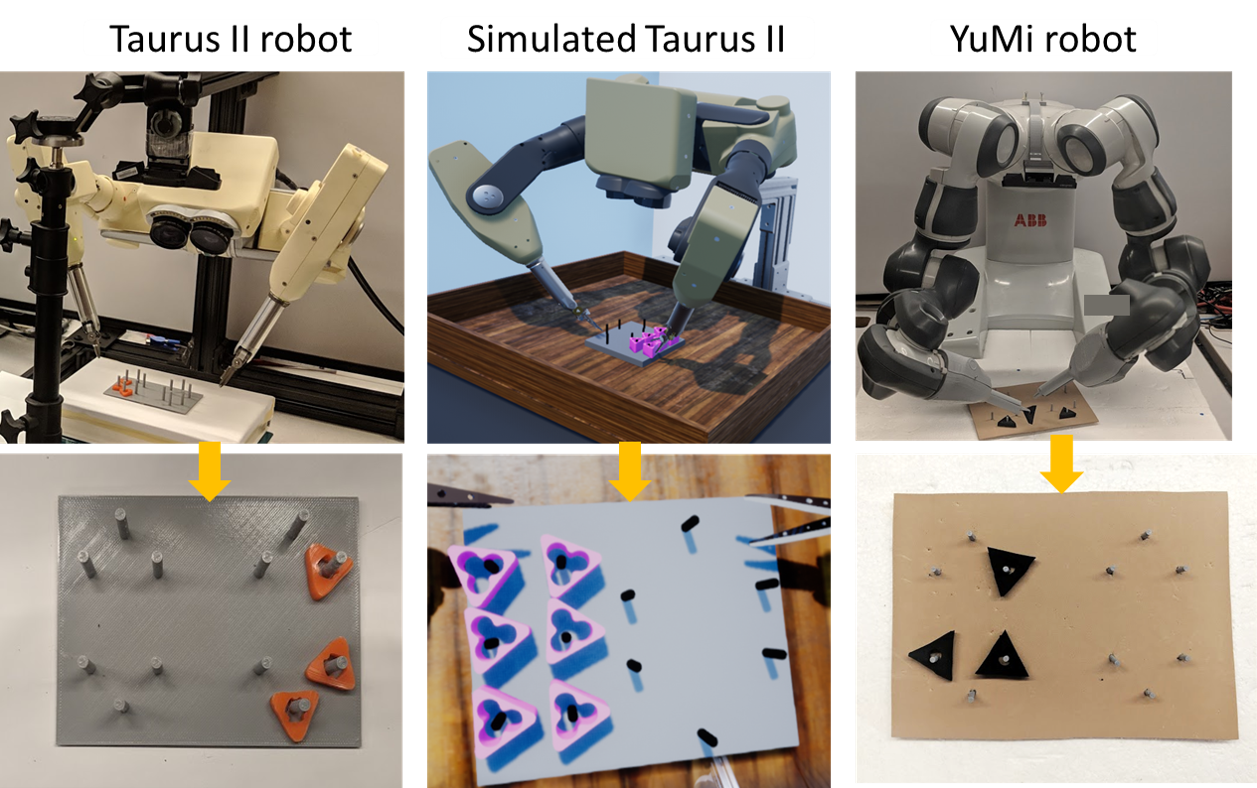}
\caption{\small Experimental setup for data collection on three different robotic domains.}
\label{fig:robots}
\end{figure}

The dataset contains a library of surgical motions for the peg transfer task using the Taurus II robot from SRI, a simulated Taurus II robot, and the YuMi robot as depicted in Figure \ref{fig:robots}. The data recordings fulfill the following requirements observed in other surgical robotic datasets \cite{gao_jhu-isi_2014}: 1) kinematic information of the robotic arms (left and right) at all captured frames, 2) RGB video capturing the task performance, 3) data annotations concerning atomic surgical actions (start frame ID, end frame ID and success or failure). Moreover, the \dataacronym provides both stereoscopic images for the simulated robot and the depth data for the real robots. 

\subsection{Peg transfer surgical task}
The peg transfer task is one of the five tasks present in the Fundamentals of Laparoscopic Surgery \cite{ritter_design_2007} and has been commonly used to train residents \cite{arain_comprehensive_2012, joseph_chopstick_2010}. The task consists of picking an object from a peg board with one robotic arm, transferring it to the other arm and positioning the object over a target peg on the opposite side of the board. These tasks require a high level of sensorimotor skill due to the small clearance between pegs and objects, and the limited maneuverability of the manipulator caused by multiple objects in the workspace. 

The peg transfer setup for the \dataacronym dataset has two sets of numbered poles (from 1 to 6), each object has to be picked from its peg with one gripper, transferred to the other gripper and placed in a specified peg on the other side. Variability is introduced in the dataset by randomizing the following elements in the setup: 1) Initial and final positions of objects, 2) direction of the transfer (objects on the left side are transferred to the right and vice versa) and 3) position/orientation of the peg board. Since multiple pegs were present in each trial, the order of the pegs to be transferred was selected by the user.

\subsection{Data description}
\begin{figure}[htpb]
\includegraphics[width = 1\columnwidth]{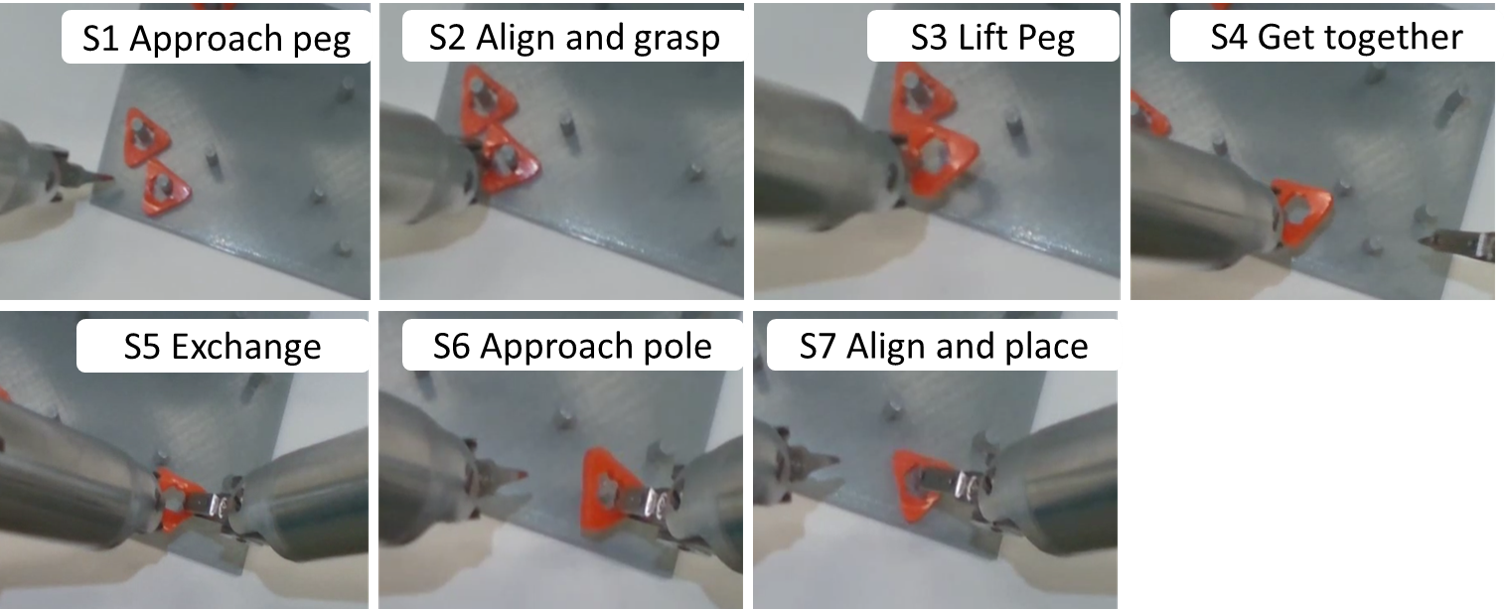}
\caption{\small Surgemes in the peg transfer task for the Taurus II robot.}
\label{fig:surgemes}
\end{figure}

The kinematic data, the RGB video and the depth video 
are segmented according to surgical gestures 
(surgemes) observed in RGB video frames. A graphical tool
was developed to facilitate the surgeme annotation based on 
RGB video recordings. A total of a total of 7 surgemes were
annotated for the peg transfer task. In addition, 
each surgeme was marked  as a success or a failure. 
Table \ref{table_surgemes} and Figure \ref{fig:surgemes} show
the list of annotated surgemes for the \dataacronym dataset.
Each peg transfer  video was associated with an annotation file that describes the following for each surgeme:  name, the start and end frame, and whether the surgeme execution was a 
success (True) or a failure (False). 
In addition, timestamps were stored
for each recording which allow to synchronize all the
recordings (depth, kinematic and controller data) with respect to the RGB videos. 

 
\begin{table}[ht]
\caption{\small Surgical gestures in the peg transfer task. The columns indicate surgeme ID, name of the surgeme, number of instances present for each surgeme for the simulator, real Taurus and the YuMi robot. }
\label{table_surgemes}
\begin{center}
\begin{tabular}{c|c|c|c|c}
\hline
ID & Surgeme name & \# Sim & \# Taurus & \# YuMi \\
\hline
\hline
S1 & Approach peg & 192 & 111 & 117\\
\hline
S2 & Align \& grasp & 206 & 115 & 123 \\
\hline
S3 & Lift peg & 203 & 112 & 123\\
\hline
S4 & Transfer peg - Get together & 180 & 113 & 117\\
\hline
S5 & Transfer peg - Exchange & 175 & 113 & 118\\
\hline
S6 & Approach pole & 167 & 109 & 117\\
\hline
S7 & Align \& place & 163 & 107 & 116 \\
\hline
\end{tabular}
\end{center}
\end{table}

\subsection{Data collection with the Taurus robot}\label{taurusdatacollection}
The Taurus II robot has two 7 degrees of freedom (DOF) arms that are controlled using the Razer Hydra\textsuperscript{\textregistered} over a stereoscopic display. The Taurus robot is interfaced with two foot pedals that allow to switch control between the arms and the camera. A \textit{clutch} pedal was used to toggle the robot between operation mode and standby mode to enable the user to reset to an ergonomic configuration for manipulation. In this setup, the robot only moves when the \textit{clutch} of the foot pedal is pressed.

The \dataacronym dataset for the Taurus robot consists of a total of 108 peg transfers, collected for 3 subjects over 36 trials, each trial included 3 randomized peg transfers. Variability is introduced in the data by randomizing three elements per trial: 1) the \textit{tilt} of the peg board (left-tilt, no-tilt, right-tilt), 2) the \textit{position} of the objects and 3) the \textit{starting point} of the peg transfer task (objects on the right side or objects on the left side). The experiment followed a full factorial design for the \textit{starting side} and \textit{tilt} of the peg board. For the initial positions of the objects, a set of three distinct random numbers from 1 to 6 were chosen.

The data collected from the Taurus robot includes RGB video and depth video recorded from the top view Realsense\textsuperscript{\textregistered} camera. In addition, the kinematic data of the Taurus robot's end-effector is captured using 16 kinematic variables as shown in Table \ref{tab:kinematic_var}. For each arm, we recorded: the rotation matrix of the wrist (nine values), the translation in x, y and z coordinates of the wrist with respect to the robot origin and the gripper state, which is a value between 30 and 100 (30 when completely closed and 100 when the gripper is completely open).

\subsection{Data Collection with the Taurus Simulation Environment}
 The simulated Taurus robot has two 7 DOF arms that are controlled using the Oculus Rift touch controllers. Similar to the Taurus robot, the simulated Taurus system has a foot pedal that enables the motion of the robot and allows to switch control between the arms and the stereoscopic camera.

The setup for the data collection consisted of six subjects performing a total of 42 trials, where each trial had six peg transfers. Thus, a total of 252 transfers were collected. At the beginning of every trial, a set of random pickup and drop-off locations were generated for the user to follow. The peg board was rigidly attached to the ground i.e. it is at a fixed orientation in the simulator. 

This dataset includes the following recordings: the kinematic data of the robot's wrist and RGB videos recorded from the point of view of the user's virtual environment, stereo view of both the left and right robot's cameras (the two stereo videos can be used to compute the depth information). The kinematic data consists of 14 kinematic variables that represent the robot's end-effector pose, as displayed in Table \ref{tab:kinematic_var}. It also includes the wrist orientation (yaw, pitch and roll angles), the translation in x, y and z coordinates with respect to the robot's origin, the joint angles (7 DOF) and the griper state (a value between 30 and 100) for both the arms.

\subsection{Data collection with the YuMi robot}
The YuMi collaborative robot has two 7 DOF arms adapted to surgical tasks using 3D printed gripper extensions \cite{sanchez-tamayo_collaborative_2018}. The end-effectors of the robot are controlled using the HTC VIVE controllers. A peg transfer setup similar to the Taurus real robot was designed for this system and variability was introduced by randomizing the three elements described in section \ref{taurusdatacollection}. The data collection of the YuMi robot is comprised of 40 trials collected by one subject. Each trial included three peg transfers comprising a total of 120 peg transfers. For this setup, the recorded data includes RGB video and depth video obtained from the top view using a Realsense\textsuperscript{\textregistered} camera. In addition, the kinematic data of each robot arm is captured using 20 kinematic variables that provide joint state information, translation of the tooltip in x,y,z coordinates, rotation matrix of the tooltip with respect to the robot's origin, and gripper state (see Table \ref{tab:kinematic_var}).

The YuMi robot is significantly different from the Taurus robot. For instance, the Taurus II is designed specifically for small dexterous tasks such as bomb disposal and surgery, while the YuMi is suitable for larger workspaces and collaborative tasks. For this reason, the setup for peg transfer using the YuMi was scaled by a factor of 2 (the size of the peg board is larger). Other differences include robot morphology and the interface used for manipulation. The pitch, yaw, and roll angles of the Taurus are computed with respect to the wrist, which causes minimal movement of the wrist when the operator changes the orientation the tool. In contrast, the YuMi robot has a kinematic control that reorients with respect to the tooltip. Given the large distance between the tooltip and the wrist, even minor changes in the orientation of the tool causes large motions at the robot's wrist and the arm configuration.

\begin{table}[ht] 
\caption{\small Kinematic variables. Note that $ts$ is the Unix timestamp, $\vec{J}$ is the vector of joint angles, $\vec{p}$ is the position vector (x, y and z), $\vec{\theta}$ be the Euler angles (yaw, pitch and roll), $gs$ is the gripper state of the end-effector and $R$ be the 3 x 3 rotation matrix. }
\label{tab:kinematic_var}
\begin{center}
\begin{tabular}{c|c||c|c||c|c}
\hline
\multicolumn{2}{c||}{\textbf{Taurus}} &
\multicolumn{2}{c||}{\textbf{Taurus Simulator}} &
\multicolumn{2}{c}{\textbf{YuMi}} \\ \hline
ID & Variable & ID & Variable & ID & YuMi \\ \hline 
\hline
1       & $ts$  & 1  & $ts$ & 1 & $ts$\\ \hline
2-13    & $R$ and $\vec{p}$  & 2-4 & $\vec{p}$ & 2-8 & $\vec{J}$\\ \hline
        & - & 5-7 & $\vec{\theta}$ & 9-11 & $\vec{p}$\\\hline 
14-16   & $\vec{p}$ & 8-14 & $\vec{J}$ & 12-20 & $R$\\\hline
17      & $gs$ & 15 & $gs$ & 21 & $gs$\\\hline
\end{tabular}
\end{center}
\end{table}

\section{Experiments and results}
The data collected from the simulator and the real robots were in different dimensions as shown in Table \ref{tab:kinematic_var}. The first step in the pipeline was to ensure that the dimensions of the data are equivalent in the three domains. Hence, we reduced the feature dimension by considering the features that are commonly shared between these two domains (position, orientation and gripper status of the end-effector). Overall, we considered 14 features per frame (seven features in each arm). 

The experiments conducted in this work are two-fold: 1. Train and test on the data obtained from the same domain (no-transfer scenario) and 2. Train on one domain and test on the other (domain-transfer scenario). Furthermore, we considered two kinds of classification tasks: 1. frame-wise and 2. sequence-wise. The former method associates each frame to a particular class label and treats each frame as a sample point (\textit{frame-wise instances}), while the latter method considers the entire surgeme (sequence of frames) as a single sample point (\textit{sequence-wise instances}). Next, we used three supervised learning methods for our experiments: 1. Support Vector Machines (SVM), 2. Random Forest (RF) and 3. Multi-layer Perceptron (MLP). These three learning techniques are commonly used in the machine learning community for creating the baselines for classification. We used the scikit-learn \cite{scikit-learn} implementation of these models for our experiments. 

\textit{Hyperparameter setting.}  We used a linear kernel for SVM classifier. For RF, we set $n\_estimators = 200$ (number of trees in the forest), and maximum depth = 10. For the MLP, we used a hidden layer of size = 100, $tanh$ as the activation function with $adam$ as the optimizer.

Each surgeme instance consists of a variable number of frames. Hence, we re-sampled (via linear interpolation) the original instances to a fixed number of frames to generate \textit{sequence-wise} instances. Next, we concatenated the seven features corresponding to each frame, for both the arms, to create a single feature vector. 
In our case, we set the number of frames to 40 and each \textit{sequence-wise} instance is a 560 dimensional vector ($40 \times 7 \times 2$). In our experiments, we did not differentiate the surgeme instances based on how well they were performed (success or failure).

\subsection{Surgeme classification}
In the first experiment, our goal is to study the classification in the \textit{no-transfer} scenario where the learning model is trained and tested on the data coming from the same domain. In other words, the training and testing data follow the same distribution. We have two such scenarios: train and test on 1. simulator data ($S\rightarrow S$) and 2. real robot data ($R \rightarrow R$). In our experiments, we have used 60-40\% split for training and testing respectively. Furthermore, we have used five-fold cross validation approach to tune the learning parameters (or weights) of the models.

Next, we performed the classification on both the frame-wise and sequence-wise instances as shown in the Table \ref{tab:no_transfer}. The \textit{sequence-wise} features contain the temporal information embedded into them. Hence these features give superior accuracy except for the random forest approach on simulator data. Therefore, we used \textit{sequence-wise} features for the experiments associated with the domain transfer. We conducted the two-sided paired t-test and found that SVM significantly outperforms the RF and MLP with the \textit{sequence-wise} features ($p < 0.05$).

\begin{table}[h]
\caption{\small Classification accuracy on the no-transfer scenario for both the frame-wise and sequence-wise features. $S$ is the Taurus simulator, $R_1$ is Taurus robot, and $R_2$ is the YuMi robot.}
\label{tab:no_transfer}
\begin{center}
\begin{tabular}{c|c|c|c||c|c|c}
\hline
\multicolumn{4}{c||}{\textbf{Sequence-wise}} & \multicolumn{3}{c}{\textbf{Frame-wise}} \\ \hline
\hline
 & RF & SVM & MLP & RF & SVM & MLP \\ \hline 
$S \rightarrow S$ & $88 \pm 2$ & $87 \pm 1$ & $78 \pm 4$ & $86 \pm 0$ & $58 \pm 1$ & $73 \pm 1$ \\ \hline
$R_1 \rightarrow R_1$ & $94 \pm 2$ & $92 \pm 1$ & $92 \pm 2$ & $95 \pm 0$ & $60 \pm 0$ & $92 \pm 1$ \\ \hline
$R_2 \rightarrow R_2$ & $91 \pm 1$ & $93 \pm 1$ & $95 \pm 1$ & $88 \pm 1$ & $48 \pm 1$ & $86 \pm 1$ \\ \hline 
\hline
\end{tabular}
\end{center}
\end{table}

\subsection{Domain transfer: Simulator to real robot}
In this experiment, our goal is to build a learning model that was trained on the data obtained from one domain but tested on the data coming from a new domain. In other words, the input data distribution is different in the train and test datasets. We start with the assumption that it is much easier to collect the data from the simulator when compared to the data from the physical surgical robot. Therefore, the real robot's data is assumed to be limited or not available in the extreme cases. Hence, we trained our models on the simulator data with a very little or no data from the real robot and test this model on the real robot's data. 

To simulate the limited availability of the real data, we added a small fraction ($\alpha$) of the real data into the training, where $\alpha$ was varied from zero to one. The value of $\alpha$ is defined as the ratio of number of examples of the real data to the simulator data present in the training. An $\alpha = 0$ indicates \textit{complete transfer}, where there is no real data present in the training, while $\alpha = 1$ implies that the data from the simulator and real data are in equal proportions. 

\begin{figure}[h]
\centering
\includegraphics[width=1\columnwidth]{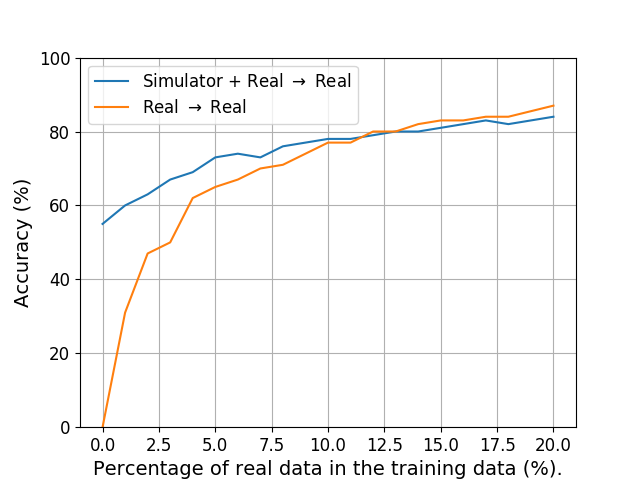}
\caption{\small \small Performance comparison for transfer learning from Taurus simulator to Taurus Robot using SVM. The blue curve indicates the transfer accuracy and the blue curve indicates accuracy on real data without transfer.}
\centering
\label{fig:transfer-abs}
\end{figure}

\begin{table}[h]
\caption{ \small Domain transfer accuracy when the models are trained on the domain $C$ but tested on domain $R$. Note that the domain $R$ is the real Taurus domain $C$ is the combination of $S$ and $R$.}
\label{tab:transfer}
\begin{center}
\begin{tabular}{c|c|c|c|c|c|c|c}
\hline
\multicolumn{2}{c|}{} & \multicolumn{2}{c|}{RF} & \multicolumn{2}{c|}{SVM} & \multicolumn{2}{c}{MLP} \\ \hline 
$\alpha$ \% & \# & $C \rightarrow R$ & $R \rightarrow R$ & $C \rightarrow R$ & $R \rightarrow R$ & $C \rightarrow R$ & $R \rightarrow R$ \\ \hline 
\hline
0 & 0 & $34 \pm 3$ & $0 \pm 0$ & $55 \pm 2$ & $0 \pm 0$ & $40 \pm 2$ & $0 \pm 0$ \\ \hline
3 & 24 & $53 \pm 3$ & $39 \pm 4$ & $67 \pm 2$ & $50 \pm 5$ & $57 \pm 3$ & $51 \pm 4$ \\ \hline
6 & 49 & $66 \pm 1$ & $64 \pm 4$ & $74 \pm 0$ & $67 \pm 3$ & $68 \pm 4$ & $72 \pm 2$ \\ \hline
9 & 73 & $77 \pm 3$ & $73 \pm 5$ & $77 \pm 1$ & $74 \pm 3$ & $74 \pm 1$ & $80 \pm 3$ \\ \hline
12 & 98 & $78 \pm 1$ & $80 \pm 3$ & $79 \pm 2$ & $80 \pm 1$ & $78 \pm 1$ & $85 \pm 1$ \\ \hline
15 & 123 & $84 \pm 2$ & $84 \pm 2$ & $81 \pm 1$ & $83 \pm 0$ & $82 \pm 2$ & $86 \pm 1$ \\ \hline
18 & 147 & $85 \pm 0$ & $85 \pm 0$ & $82 \pm 1$ & $84 \pm 1$ & $79 \pm 3$ & $88 \pm 1$ \\ \hline
\hline
\end{tabular}
\end{center}
\end{table}

Moreover, \textit{accuracy} ($\gamma$ - the percentage of test examples that are correctly classified) and confusion matrices are used as performance metrics to evaluate how well the model is performing on the unseen data. Let us define $\gamma^M_{A\rightarrow B}$ as the test accuracy of the model $M$ trained on the data from domain A and tested on the data from domain B. Note that $\gamma^M_{A\rightarrow B}$ depends on the value of $\alpha$. Also, let us define $C$ as the domain that is a combination of both real and simulator domains. The goal of this study is to determine the behavior of $\gamma^M_{C\rightarrow R}$ (trained on both simulator and real data and tested on the real data) and $\gamma^M_{R\rightarrow R}$ (trained and test on real data) as the value of $\alpha$ increases. 

Figure \ref{fig:transfer-abs} shows the behavior of $\gamma^{SVM}_{R\rightarrow R}(\alpha)$ and $\gamma^{SVM}_{C\rightarrow R}(\alpha)$ for SVM classifier. In the extreme case when there is no real data in the training process (i.e. $\alpha = 0$ - \textit{complete transfer}), the accuracy is approximately 55\%. As the value of $\alpha$ gradually increases from $0$ to $0.2$, the value of $\gamma^{SVM}_{C\rightarrow R}$ increases and converges in comparison to $\gamma^{SVM}_{R\rightarrow R}$. At $\alpha = 0.18$, the values of $\gamma^{RF}_{C\rightarrow R}$ and $\gamma^{RF}_{R\rightarrow R}$ are approximately equal to 80\%. Moreover, this plot shows that adding the simulator data into the training procedure helps improve the accuracies considerably on the real data.

\begin{figure}[h]
\centering
\includegraphics[width=1\columnwidth]{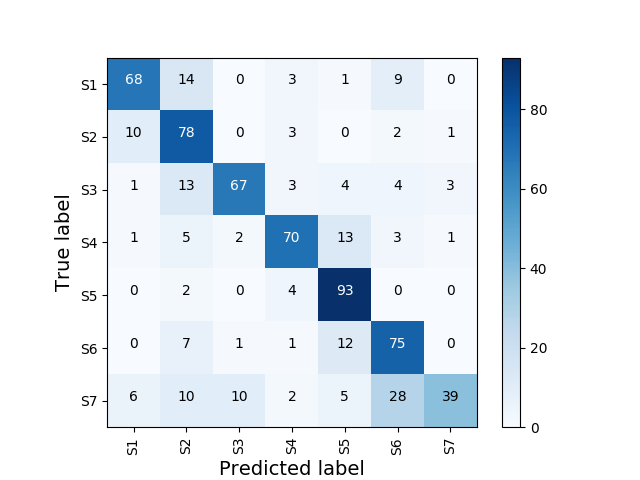} 
\caption{\small Confusion matrix for transfer learning with SVM using 5\% real Taurus robot data.}
\centering
\label{fig:conf-mat}
\end{figure}

Table \ref{tab:transfer} shows the transfer accuracies obtained using three learning models for a range of values of $\alpha$. Note that there are 1286 samples in the simulator domain. The second column (\#) is the number of examples of the real data present in the training procedure. In the first row, it was assumed that there were no real data examples in the training, hence accuracy in $R\rightarrow R$ is 0. The confusion matrix presented in Figure \ref{fig:conf-mat} shows the transfer accuracies for all the classes at $\alpha = 0.05$. Overall, the Table \ref{tab:transfer} shows that it is beneficial to augment the training data with the samples obtained from robotic simulators. Irrespective of the learning model, it helps greatly enhance the performance of those models on the real robot's data.

\begin{figure}[h]
\centering
\includegraphics[width=1\columnwidth]{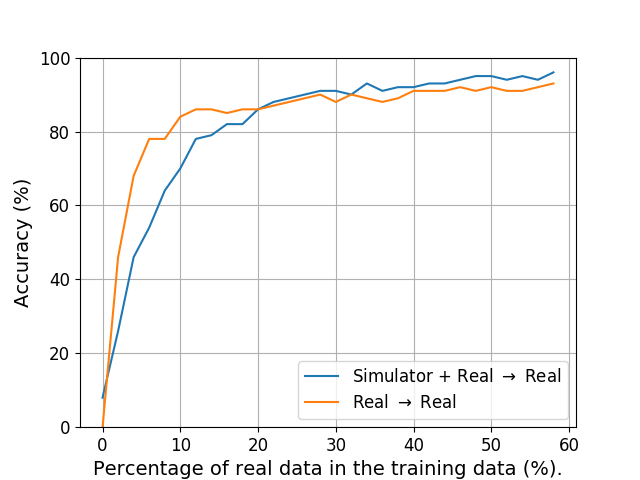}
\caption{\small Performance comparison for transfer learning from the Taurus simulator to the YuMi Robot.}
\centering
\label{fig:transfer-yumi}
\end{figure}

In the last experiment, our goal is to verify the transfer across the robots i.e. to train the model on the simulation data of Taurus robot and test on the data of YuMi robot. The workspace, morphology and dimension of the data of these robots are significantly different. Hence we used the principal component analysis (PCA) to rank the features based on the Eigen values and transform the data into the Eigen space. Next, we chose a fixed number of features (a value of 8 is used in our experiments) in the Eigen space in order to unify the feature dimensions. We followed a similar experimental design to test the transfer of knowledge from Taurus simulator to YuMi domain. The accuracy of \textit{complete transfer} scenario is close to the random accuracy indicating that the model is overfitting to the simulator as there is no data from the real robot in the training. However, a slight increase in the number of examples of YuMi data in the training, the transfer accuracies swiftly increase. For instance, when the number of YuMi examples = 100, the transfer accuracy is approximately 80\% as shown in the Figure \ref{fig:transfer-yumi}. When we add more than 200 examples of YuMi in the training, the transfer accuracies marginally surpasses the accuracies in $R\rightarrow R$ scenario. This shows that the transferring the knowledge between two completely different robots is a challenging task and requires relatively more examples to achieve superior transfer accuracies.

\section{Discussion and future work}

The transfer learning task considered in this paper is an example of \textit{domain adaptation} where the data distribution of the source and target domains are considerably different while the class labels remain the same. In our experiments, the source domain is nothing but the simulator where a large amount of data is available whereas the target domain is the real Taurus robot. The interaction modality of the Taurus robot and the simulated robot are completely dissimilar. Given that the configuration of the real Taurus robot and the simulator is similar, transferring the knowledge to the physical domain is relatively easier in the case of the real Taurus robot in comparison to the YuMi robot. Hence the \textit{transfer accuracy} is significantly higher for the real Taurus. Furthermore, the \textit{transfer accuracy} obtained on the YuMi robot increases with the increase in $\alpha$ and surpasses $R\rightarrow R$ scenario for $\alpha > 0.2$. In other words, the amount of real data needed to achieve the \textit{transfer accuracy} of 80\% is much higher for the YuMi ($\alpha = 0.3$) in comparison to real Taurus ($\alpha = 0.12$). The relatively lower transfer accuracies obtained on YuMi data show that the transfer from one robot to another robot is a challenging task.

It is worth revisiting the process of creating the \textit{sequence-wise} features. Given that the surgeme instances are of different length, we used linear interpolation to ensure that the instances are of uniform length. This method of interpolation, elongates the surgemes that are shorter and compresses the surgemes that are longer. It is interesting to study the effect of re-sampling on the \textit{transfer accuracy}. Nevertheless, the issues associated with re-sampling can be eliminated by using the deep models that are specialized for the problems with non-uniform features lengths such as Long-Short-Term-Memory models or conditional random fields. 

Furthermore, we mentioned in the results that the \textit{sequence-wise} features provide significantly better accuracies in comparison to the \textit{frame-wise} features for the transfer learning tasks. Note that the sequence classification approach assumes that the test data is annotated beforehand with respect to the start and end of the surgemes. However, this assumption is not valid when we want to deploy these trained classifiers in real-time as the real-time data cannot be segmented beforehand. In contrast, the \textit{frame-wise} classification approach does not require the surgemes to be segmented. Hence the \textit{frame-wise} classifier can potentially act as the model that can be used to know the start and the ending point of the surgemes. 

In this regard, we have made our database publicly available to encourage researchers to further investigate this problem of transfer learning between the robots or from simulator to a real robot. In addition to the RGB-D videos and the kinematic data, we also provide bounding boxes of the objects and pegs for each frame. These annotations are created in a semi-autonomous manner i.e. first, the color-based image processing techniques were used to create the bounding boxes automatically and next, a human annotator was asked to verify those annotations and manually annotate the frames with erroneous annotations. The instructions to obtain the data are available at \href{https://github.com/nmadapan/Forward_Project.git}{https://github.com/nmadapan/Forward\_Project.git}.

The potential future works and applications of our dataset include: 1. Incorporating the object bounding boxes and visual data (RGB-D images) into the feature vectors to improve the transfer accuracies, 2. Developing task specific deep models with the goal of transferring the knowledge from one robot to the other, 3. Learn to predict the surgemes from one physical robot instead of the simulator and test on an other physical robot (in our case, train on YuMi and test on Taurus), 4. Learning to predict the surgemes only with the partial information, and 5. Developing the learning models that do not require the surgemes to be segmented beforehand. 







\section{Conclusions}
The main goal of this paper is to learn to transfer the knowledge (surgical skills in our case) from one domain (either simulator or a physical robot) to another domain (physical robot). Previous datasets concerned with surgical tasks were mainly focused on a unique robotic platform (e.g. da-Vinci). This limits the researchers to explore novel ways to transfer the surgical skills learned from one platform to the other. Therefore, it is essential to have the data collected from various robotic platforms. Hence, we created a database \dataname of surgical robotic skills (peg transfer task) collected from three domains: simulated Taurus robot, real Taurus robot and YuMi industrial robot. In addition to developing the dataset, this paper proposes a, simple yet effective, transfer learning methodology to improve the learning for a surgical gesture classification task over real robot's data using the data obtained from the simulation. We conducted an experiment on the dataset and presented three supervised models as baselines for surgeme classification. Results show that augmenting the training data with simulator data considerably improves the accuracies of prediction on the real data. More specifically, in the extreme case when there is no real Taurus robot's data present in the training, the \textit{transfer accuracy} on the real Taurus data is 55\%. Furthermore, the \textit{transfer accuracy} increases significantly as the real robot's data is added to the training process. 



\section*{Acknowledgments}
This work was supported by both the Office of the Assistant Secretary of Defense for Health Affairs under Award No.  W81XWH-18-1-0769. Opinions, interpretations, conclusions and recommendations are those of the author and are not necessarily endorsed by the funders.





\balance

\bibliography{mybib}{}
\bibliographystyle{ieeetr}
\addtolength{\textheight}{-3cm}   


\end{document}